\documentclass[10pt,twocolumn,letterpaper]{article}

\usepackage{cvpr}             
\usepackage{graphicx}
\usepackage{amsmath}
\usepackage{amssymb}
\usepackage{booktabs}
\usepackage{mathtools, cuted}
\usepackage{mathrsfs}
\usepackage{breqn}

\usepackage{array}
\usepackage{amsfonts}
\usepackage[margin=1.0in]{geometry}
\usepackage{tabularx}
\usepackage{multirow}
\usepackage{url}
\usepackage{makecell}
\newcolumntype{P}[1]{>{\centering\arraybackslash}p{#1}}
\usepackage{caption}
\usepackage{subcaption}
\usepackage[pagebackref,breaklinks,colorlinks]{hyperref}
\usepackage[capitalize]{cleveref}
\crefname{section}{Sec.}{Secs.}
\Crefname{section}{Section}{Sections}
\Crefname{table}{Table}{Tables}
\crefname{table}{Tab.}{Tabs.}

\begin{document}

\title{Improving Axial-Attention Network Classification \\via Cross-Channel Weight Sharing\thanks{Partially supported by 
an Alfred and Helen M.\ Lamson  endowed chair award.}}

\author{Nazmul Shahadat, Anthony S.\ Maida\\
University of Louisiana at Lafayette\\
Lafayette LA 70504, USA\\
{ nazmul.ruet@gmail.com, maida@louisiana.edu}}
\maketitle

\begin{abstract}
In recent years,
hypercomplex inspired neural networks (HCNNs) have been used to improve deep learning architectures due to their ability to enable channel-based weight sharing, treat colors as a single entity,
and improve representational coherence within the layers.
The work described herein studies the effect of replacing existing layers in
an Axial Attention network with their representationally coherent variants
to assess the effect on image classification.
We experiment with the stem of the network, the bottleneck layers, and the
fully connected backend, by replacing them with representationally coherent
variants.
These various modifications lead to novel architectures which all yield improved accuracy
performance on the ImageNet300k classification dataset. 
Our baseline networks for comparison were the original real-valued ResNet, the original quaternion-valued
ResNet, and the Axial Attention ResNet.
Since improvement was observed regardless of which part of the network was modified,
there is a promise that this technique may be generally useful in improving
classification accuracy for a large class
of networks.
\end{abstract}

\section{Introduction}
This work studies the effect of adding representationally coherent layers
to axial-attention networks to improve image classification accuracy.
In this work, a representationally coherent layer means a layer with the ability to 
discover and represent cross-channel correlations in its inputs. This functionality is implemented using calculations inspired by hypercomplex number systems such as quaternions \cite{gaudet2018}, vector maps \cite{Gaudet2021a}, and 
parameterized hypercomplex multiplication (PHM) \cite{zhang2021beyond}. And, a representationally coherent network is a network where better representationally feature maps, generated by any HCNN, are used.

A core example illustrating the utility of representational coherence comes
from auto-encoder networks which can be trained to reconstruct color from
grayscale input. \cite{parcollet2019} showed that the channel-based weight-sharing property of a quaternion-valued auto-encoder allowed the discovery of input correlations that supported the reconstruction
of color from grayscale images, and this was not possible using a real-valued 
auto-encoder. This result was subsequently confirmed using a vector map auto-encoder that used a similar form of weight sharing.

We hypothesize that this improved ability to capture relationships between input channels may apply to other multidimensional inputs besides color.
It may depend on whether the input data contains interwoven cross-channel relationships
to capture.
We will say that layers that use this form of hypercomplex inspired weight sharing are 
\textit{representationally coherent}.

Our experiments modified Axial Attention networks \cite{ho2019axial,wang2020axialdeeplab} by replacing existing layers with 
representationally coherent layers in different parts of the network.
These modifications included:

\begin{enumerate}
\item Modifying the bottleneck blocks with representationally coherent blocks.
\item Replacing the fully connected, real-valued backend with a 4D parameterized hypercomplex
      multiplication (PHM).
\item Replacing the real-valued convolutional stem (frontend) with a quaternion-valued convolutional
      stem.
\end{enumerate}

We varied the depths of the original axial attention networks by using 26, 35, and 50 layers. In addition to the axial attention networks, we compared our results to real-valued, and quaternion-valued ResNets. We found, in general, that adding representationally coherent layers improved performance over the ResNets, quaternion ResNets, and the axial-attention networks that were tested. Also, the most notable improvement occurred by adding representational coherence to the backend in the form of a 4D PHM layer.

\section{Rationale for the Proposed Method}
\label{sect_mainHypothesis}
The main hypothesis of this paper is that multichannel feature map representations
that are used as input to attention modules can be modified to improve their
effectiveness. Our approach is to use the output of 
quaternion modules to provide improved input representations
to the attention modules. The rationale follows.

\cite{parcollet2019} showed that a quaternion-valued, auto-encoder can be trained 
to reconstruct color from grayscale input images whereas a real-valued autoencoder cannot.
Thus, a trained quaternion-valued layer generates a richer representation because it allows implicit relationships among the color channels of data and these relationships cannot be captured in a comparable real-valued auto-encoder. \cite{parcollet2019} attribute this functionality to a weight-sharing property found in the Hamiltonian product (quaternion multiplication) that is used to implement convolution. These are not happened for real-valued networks.
The representation produced by the quaternion network captures more information about the interrelationships between the color input channels, and in this sense, we can say that it is a richer 
\textit{interwoven} or
\textit{interlinked representation}
of the information contained in the input channels.

We hypothesize that this improved ability to capture relationships between input channels is likely to apply to other multidimensional inputs besides color \cite{gaudet2020generalizingarXiv}. It may depend on whether the input data happens to contain \textit{interwoven cross-channel relationships} of this kind to capture. 
In the remainder of this paper, we use the term \textit{interwoven/interlinked representation}
or \textit{feature map} to refer to the output of a quaternion layer.

\section{Background and Related Work}

\subsection{Quaternion Convolution}
\label{subsection_quatconv}
A useful introduction to quaternion algebra and neural networks appears in \cite{parcolletsurvey2020}.

The paper \cite{trabelsi2018deep} extended the principles of convolution,
batch normalization, and weight initialization from real-valued networks to
complex-valued networks.
\cite{gaudet2018}, in turn, extended these principles to quaternion-valued convolutional
networks (QCNNs). For implementation purposes,
quaternion calculations can be decomposed into operations on 4-tuples of real numbers
and quaternion operations can be implemented on these 4-tuples. The take home message from this is that a quaternion convolution module must accept four channels of input. A quaternion layer can accept more than four input channels, say $m$, as long as $m$ a multiple of four. In this case, the layer must hold $m/4$ separate quaternion convolution modules, each with their own weight sets.

The paper \cite{parcollet2019} analyzed the Hamilton product, which underlies
the quaternion convolution, to identify the 
mechanism for improved performance in quaternion models.
Let us denote the Hamilton product of I and Q as

\begin{equation}
O_q = I_q \circledast W_q,
\label{HamiltonProduct1}
\end{equation}

\noindent
where the subscripts $q$ indicate that I, W, and Q are quaternion numbers.
When expanded into real-valued 4-tuples, this can be viewed as a linear mapping from
a neural layer of four units, representing $I_q$, to another layer of four units
representing $O_q$. However, 
Instead of using 16 independent weights to connect the layers, the four weights in
$W_q$ are repeatedly substituted within the 16 weights, so the mapping only uses
four independent weights.
This weight sharing forces the model to learn cross-channel interrelationships 
in the data.
The paper \cite{parcollet2019} confirmed this experimentally by comparing 
the ability of a real-valued
autoencoder with a quaternion-valued autoencoder
on the task of reconstructing  color images.
The quaternion network learned to accurately reconstruct the colors in unseen
images but the real-valued network did not. More recently, the paper \cite{gaudet2020generalizingarXiv,Gaudet2021a} showed that the properties
of the quaternion convolution were due entirely to weight sharing and had no
dependence on the quaternion algebra.
The quaternion convolutions in our model are 1x1, so they can be interpreted as fully connected layers
with shared weights.

\begin{figure*}
    \centering
     \begin{subfigure}[b]{0.2\textwidth}
        \centering
        \includegraphics[width=0.85\textwidth,height=200pt]{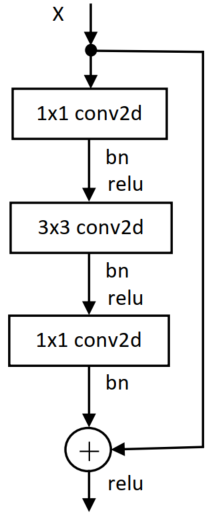}
        \caption{ResNet}
        \label{ResNet}
    \end{subfigure}
    \hfill
    \begin{subfigure}[b]{0.2\textwidth}
        \centering
        \includegraphics[width=0.85\textwidth,height=200pt]{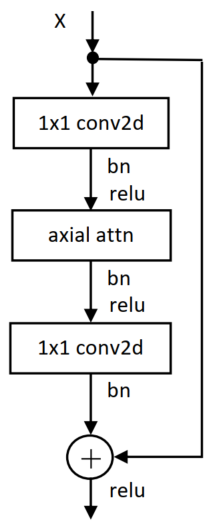}
        \caption{Axial ResNet}
        \label{Axial ResNet}
    \end{subfigure}
    \hfill
    \begin{subfigure}[b]{0.27\textwidth}
        \centering
        \includegraphics[width=0.8\textwidth,height=220pt]{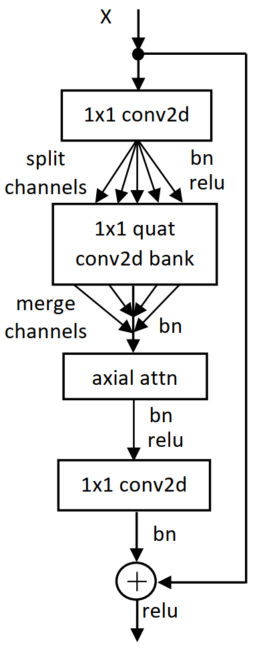}
        \caption{QuatE}
        \label{Quaternion Enhanced Axial ResNet}
    \end{subfigure}
    \hfill
    \begin{subfigure}[b]{0.23\textwidth}
        \centering
        \includegraphics[width=\textwidth,height=200pt]{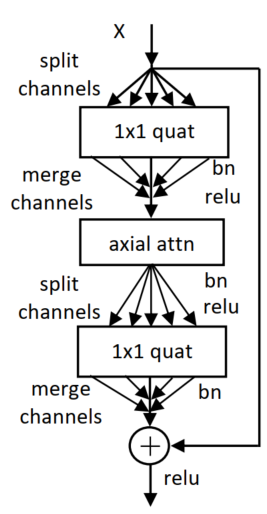}
        \caption{RepA}
        \label{RepAA}
    \end{subfigure}
    \hfill
\caption{Bottleneck types. ``bn'', ``attn'', and ``quat'' stand for batch normalization, attention, and quaternion, respectively. (a and b) Original Bottleneck modules found in ResNet \cite{heresnet2016}, and Axial-ResNet \cite{wang2020axialdeeplab}, respectively. (c) Quaternion enhanced (QuatE) axial-ResNet bottleneck block (our proposed model), and (d) Quaternion based representational Axial-ResNet bottleneck module (RepAA) used in our model.}
    \label{fig:proposedWithAttention}
\end{figure*}

\subsection{PHM Layer}
\label{subsection_PHM}
This section explains 4D generalized hypercomplex networks for fully-connected layer,
using parameterized hypercomplex multiplication (PHM) \cite{zhang2021beyond}. 
The 4D (quaternion) PHM (QPHM) layer works in a similar way of the quaternion networks. The Kronecker product is used to construct the parameter matrix. 
The PHM based fully connected hypercomplex transformation which transforms the input $\mathbf{x}\in  \mathbb{R}^d$ into an output $\mathbf{y}\in \mathbb{R}^k$, is defined as $ y = Hx + b$ where, $H \in  \mathbb{R}^{k \times d}$ represents the PHM layer. H is calculated as $
H = \sum_{i=1}^n \mathbf{I}_i \otimes \mathbf{A}_i$ where   $\mathbf{I}_i\in\mathbb{R}^{n\times n}$
and $\mathbf{A}_i \in \mathbb{R}^{k/n\times d/n}$ are parameter matrices and $i = 1\ldots n$ ($n=4$).
Parameter reduction comes from reusing matrices $\mathbf{I}$ and $\mathbf{A}$ in the PHM layer. 
The $\otimes$ is the Kronecker product. 
The inputs are split as $Q_{in} = Q_r + Q_x + Q_y + Q_z$ and the outputs are merged into $Q_{out}$ as $Q_{out} = Q_{ro} + Q_{xo} + Q_{yo} + Q_{zo}$ for 4D hypercomplex which are graphically explained in Figure \ref{fig:QARNet}. The 4D hypercomplex parameter matrix is discussed in \cite{zhang2021beyond} which expresses the Hamiltonian product for 4D by preserving all PHM layer properties.

\subsection{Axial-Attention Networks}
As stated in the introduction, standalone attention networks consisting of stacked
attention modules can learn to outperform deep CNNs and hybrid CNN/attention models for image classification.
Their advantage stems from their ability to detect similarities, or affinities, between pixels that
have large spatial separation in the image.
The drawback of directly using self-attention is that it is impractically computationally 
expensive, consuming $\mathcal{O}(N^2)$ resources for an image of length $N$ (using a flattened
pixel set)
and using a global window
to compare any pair of pixels in the image.
For a 2D image of height, $h$, and width, $w$, where
$N=hw$, so
the cost is $\mathcal{O}((hw)^2)=\mathcal{O}(h^2 w^2)$ to detect
similarities for any pair of pixels in the image
\cite{ho2019axial,wang2020axialdeeplab}.

Axial-attention networks reduce the cost of computing attention by decomposing the problem into consecutive 1D operations.
They were introduced in \cite{ho2019axial} for generative modeling in auto-regressive models
and use the assumption that images are approximately square so that $h$ and $w$ are both 
much less than the total pixel count, $hw$.
For simplicity, assume a square 2D image where $h = w$, so $w^2 = N$.
Axial attention only operates on one dimension at a time.
It is first applied to, say, the $h$ axis and then to the $w$ axis.
When applied to the $h$ axis, then $h=w$ attention calculations are applied to a 1D region
of length $h$.
Axial attention applied to the columns, $h$, performs $w$ self-attention operations on each column
whose total cost is $\mathcal{O}(h\cdot h^2) = \mathcal{O}(\sqrt{N}\cdot N)$.
This bound is the same for using an axial column module followed by an axial row module.

The paper
\cite{wang2020axialdeeplab} incorporated axial attention 
into a ResNet architecture \cite{heresnet2016}
to develop a model called Axial-ResNet, which
reduced the
computational requirements (described above) of the original standalone attention 
networks \cite{ramachandranstandalone2019a,hanhu2019a} used for image classification.
The conversion from ResNet to Axial-ResNet was based entirely on modifying the
bottleneck blocks within ResNet.
Figure~\ref{fig:proposedWithAttention}(a) shows the bottleneck block used in the original convolution-based ResNet.
Figure~\ref{fig:proposedWithAttention}(b) shows the axial-bottleneck block used in Axial-ResNet.
The 3x3 2D convolution used in the original ResNet is replaced by 
an axial attention module (consisting of two 1D layers)
that processes the $h$ axis
followed by the $w$ axis.

\section{Our Model: Axial-Attention with Quaternion Coherent Layers}
\label{sec:QARNetArchi} 

Our model is an axial-attention model used in \cite{wang2020axialdeeplab} and described above but modified to supply quaternion input representations to the axial attention blocks.

\begin{figure*}
    \centering
    \includegraphics[width=.95\linewidth]{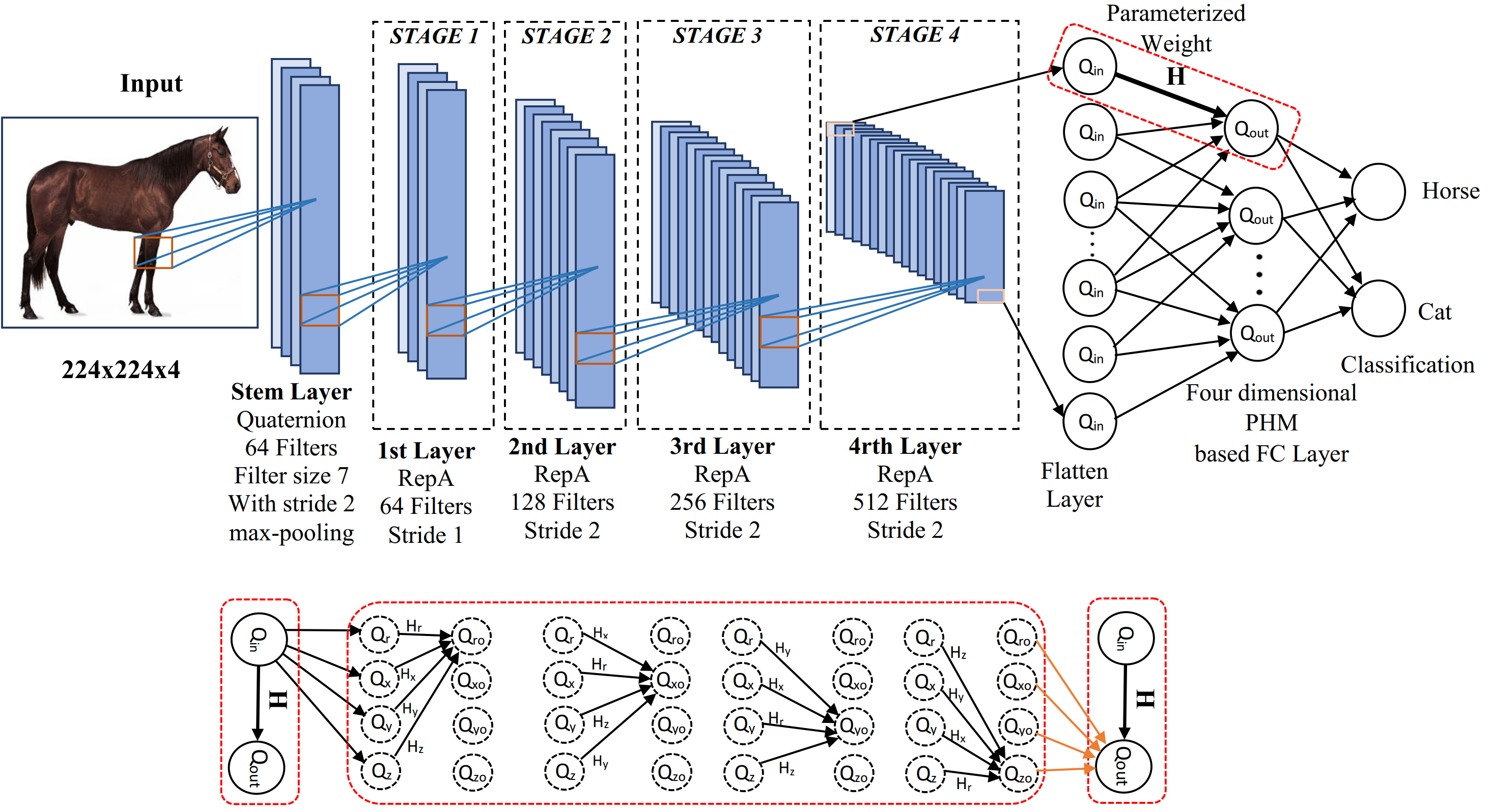}
    \caption{Proposed full representational axial-attention network with QPHM based fully-connected layer. RepA stands for quaternion based axial bottleneck block (shown in figure \ref{fig:proposedWithAttention} (right)). Here, $Q_{in} = Q_r+Q_x+Q_y+Q_z$, $H = H_r+H_x+H_y+H_z$, and $Q_{out} = Q_{ro}+Q_{xo}+Q_{yo}+Q_{zo}$ are the input, hypercomplex parameterized weight, and output respectively. The calculation of $H$ is explained in ``PHM Layer'' section \ref{subsection_PHM}.}
    \label{fig:QARNet}
\end{figure*}

Figure~\ref{fig:proposedWithAttention}(c) shows our first modification to the axial-bottleneck block which is shown in Figure~\ref{fig:proposedWithAttention}(b). It inserts a bank of 1x1 quaternion conv2d modules in front of the axial-attention module. Our proposed model is created by modifying the axial-bottleneck module, it is possible to
add a quaternion front end to the axial attention module. The purpose of the front end is to generate potentially more useful interwoven/interlinked input representations for use by the axial-attention modules. The number of input channels to the quaternion bank is constrained to be a multiple of four. Hence, 
the output channels of the top 1x1 conv2d module are split into groups of four. One quaternion 2D convolution is applied to each group of four channels.
Each quaternion convolution accepts four channels of input and produces four channels of output. Thus, the weight-sharing is compartmentalized to groups of four input channels. The set of output channels is merged (stacked) so that the number of input channels to the axial-attention module is unchanged from the original axial-attention block. In all other ways, the structure of our quaternion-modified model is identical to the axial-ResNet model.

Figure~\ref{fig:proposedWithAttention}(d) shows our final modification to the axial-bottleneck block. 
We redesign the axial-bottleneck block by removing a bank of $1 \times 1$ quaternion conv2d modules from our first proposed network (Figure \ref{fig:proposedWithAttention}(c)) and replacing $1 \times1$ convolutional down-sampling and up-sampling modules with a bank of $1 \times 1$ quaternion conv2d modules. 
The set of output channels of down-sampled $1\times 1$ quaternion is merged into input to the axial-attention modules and the output channels of axial-attention modules are split into groups of four again for $1\times1$ up-sampled quaternion layer. One quaternion 2D convolution is applied to each group of four channels. Thus, the weight-sharing is compartmentalized to groups of four input channels.

For better representation, quaternion layer is also applied in the stem layer (the first layer of the network) as a quaternion based front-end layer and in the fully-connected dense layer as a quaternion (PHM layer with four dimensions) based back end layer. To make more interwoven output representation in the bottleneck block of the network, we also use quaternion layers in the  axial-bottleneck architecture which is described in Figure \ref{fig:proposedWithAttention}(d). We renamed these fully representation 
based axial-ResNets as RepAA networks which is depicted in Figure~\ref{fig:QARNet}.

\begin{table*}
\centering
\begin{tabular}{c|c|c|c|c|c} 
\hline
Layer & \shortstack{Output\\size} &  
\shortstack{Axial-ResNet\\\cite{wang2020axialdeeplab}} & \shortstack{QuatE-1\\ Axial-ResNet (ours)} & \shortstack{QuatE-2\\ Axial-ResNet (ours)} & \shortstack{RepAA\\ network (ours)}\\ \hline

\multirow{2}{*}{Stem} & \hspace{-0.7em} 112x112 & 
\multicolumn{3}{c|}{7x7C, $64s$, stride=2} & 7x7Q, $64s$, stride=2 \\ \cline{2-6}
& $56\mathrm{x}56$ & \multicolumn{4}{c}{ $3\mathrm{x}3$ max-pool, stride=2}  \\ \hline

\shortstack{Bottle.\\ group 1} &56x56& 
$ \begin{bmatrix} 1\mathrm{x}1\mathrm{C}, 64s \\ 56\mathrm{x}56\mathrm{A}, 64s \\ 1\mathrm{x}1\mathrm{C}, 128s \end{bmatrix} \times 3 $ & 
$ \begin{bmatrix} 1\mathrm{x}1\mathrm{C}, 64s \\ 1\mathrm{x}1\mathrm{Q}, 64s \\ 56\mathrm{x}56\mathrm{A}, 64s \\ 1\mathrm{x}1\mathrm{C}, 128s \end{bmatrix} \times 3 $ &
$ \begin{bmatrix} 1\mathrm{x}1\mathrm{C}, 64s \\ 1\mathrm{x}1\mathrm{Q}, 64s \\ 56\mathrm{x}56\mathrm{A}, 64s \\ 1\mathrm{x}1\mathrm{C}, 128s \end{bmatrix} \times 3 $ &
$ \begin{bmatrix} 1\mathrm{x}1\mathrm{Q},64 \\ 56\mathrm{x}56\mathrm{A}, 64s \\ 1\mathrm{x}1\mathrm{Q},128 \end{bmatrix} \times 3 $ 
\\ \hline

\shortstack{Bottle.\\group 2} & $28\mathrm{x}28$ & 
\hspace{-0.75em}$ \begin{bmatrix} 1\mathrm{x}1\mathrm{C}, 128s \\ 28\mathrm{x}28\mathrm{A}, 128s \\ 1\mathrm{x}1\mathrm{C}, 256s \end{bmatrix} \times 4 $ &
\hspace{-0.75em}$ \begin{bmatrix} 1\mathrm{x}1\mathrm{C}, 128s \\ 1\mathrm{x}1\mathrm{Q}, 128s \\ 56\mathrm{x}56\mathrm{A}, 128s \\ 1\mathrm{x}1\mathrm{C}, 256s \end{bmatrix} \times 4 $ &
\hspace{-0.75em}$ \begin{bmatrix} 1\mathrm{x}1\mathrm{C}, 128s \\ 1\mathrm{x}1\mathrm{Q}, 128s \\ 28\mathrm{x}28\mathrm{A}, 128s \\ 1\mathrm{x}1\mathrm{C}, 256s \end{bmatrix} \times 4 $
& 
\hspace{-0.75em}$\begin{bmatrix} 1\mathrm{x}1\mathrm{Q},128 \\ 28\mathrm{x}28\mathrm{A}, 128s \\ 1\mathrm{x}1\mathrm{Q}, 256s \end{bmatrix}\times 4$\\ \hline

\shortstack{Bottle.\\group 3} & $14\mathrm{x}14$ & 
\hspace{-0.75em}$ \begin{bmatrix} 1\mathrm{x}1\mathrm{C}, 256s \\ 14\mathrm{x}14\mathrm{A}, 256s \\ 1\mathrm{x}1\mathrm{C}, 512s \end{bmatrix} \times 6 $&  
\hspace{-0.75em}$ \begin{bmatrix} 1\mathrm{x}1\mathrm{C}, 256s \\ 1\mathrm{x}1\mathrm{Q}, 256s \\ 28\mathrm{x}28\mathrm{A}, 256s \\ 1\mathrm{x}1\mathrm{C}, 512s \end{bmatrix} \times 6 $ &

\hspace{-0.75em}$ \begin{bmatrix} 1\mathrm{x}1\mathrm{C}, 256s \\ 1\mathrm{x}1\mathrm{Q}, 256s \\ 14\mathrm{x}14\mathrm{A}, 256s \\ 1\mathrm{x}1\mathrm{C}, 512s \end{bmatrix} \times 6 $ & 
\hspace{-0.75em}$\begin{bmatrix} 1\mathrm{x}1\mathrm{Q},256 \\  14\mathrm{x}14\mathrm{A}, 256s \\ 1\mathrm{x}1\mathrm{Q}, 512s \end{bmatrix} \times 6 $
\\ \hline

\shortstack{Bottle.\\group 4} & $7\mathrm{x}7$ & 
$ \begin{bmatrix} 1\mathrm{x}1\mathrm{C}, 512s \\ 7\mathrm{x}7\mathrm{A}, 512s \\ 1\mathrm{x}1\mathrm{C}, 1024s \end{bmatrix} \times 3 $ & 
$ \begin{bmatrix} 1\mathrm{x}1\mathrm{C}, 512s \\ 1\mathrm{x}1\mathrm{Q}, 512s \\ 14\mathrm{x}14\mathrm{A}, 512s \\ 1\mathrm{x}1\mathrm{C}, 1024s \end{bmatrix} \times 3 $ &
$ \begin{bmatrix} 1\mathrm{x}1\mathrm{C}, 512s \\ 1\mathrm{x}1\mathrm{Q}, 512s \\ 7\mathrm{x}7\mathrm{A}, 512s \\ 1\mathrm{x}1\mathrm{C}, 1024s \end{bmatrix} \times 3 $
 & 
$ \begin{bmatrix} 1\mathrm{x}1\mathrm{Q}, 512s \\ 7\mathrm{x}7\mathrm{A}, 512s \\ 1\mathrm{x}1\mathrm{Q}, 1024s \end{bmatrix} \times 3 $\\ \hline

Output& \shortstack{$1\mathrm{x}1$\\$\mathrm{x}1000$} &\shortstack{global avg-pool, \\FC layer, softmax} &\multicolumn{3}{c}{\shortstack{global average-pool, \\1000 outputs, QPHM, softmax}} \\ \hline
\end{tabular} 
\caption{The 50-layer architectures tested:
50-layer axial-ResNet, our proposed QuatE-1 and QuatE-2 axial-ResNet (66 layers) and RepAA network. Input is a 224x224x3 color image. The number of stacked bottleneck modules are specified by multipliers. ``C'', ``Q'', and ``A'' denote CNN, QCNN, and axial-attention layers. 
Integers (e.g., 64, 128 \ldots) denote the number of output channels. The symbol, $s=0.5$, scales the channel count (width scaling factor) for all layers of the axial-ResNet models. 
\label{tab_archiTable250}
}
\end{table*}

\section{Overview of Experiments}
\label{sec:ExperimenOverview}
Our experiments studied specific novel architectures involving modifications to the Axial Attention ResNet. Our first modification replaces the bottleneck blocks of the Axial ResNet with the quaternion bottleneck blocks depicted in Fig.\ \ref{fig:proposedWithAttention}(c). This converted 26 and 50 layer Axial Attention ResNets into 33 and 66 layer quaternion enhanced Axial Attention ResNets. We refer to instances of this architecture as ``QuatE-1'', short for ``quaternion enhanced, version 1''.
Since these QuatE-1 networks were deeper as a result of using quaternion bottleneck blocks, we also
assessed the accuracy performance of a 35 layer unmodified Axial Attention network as a control.

The second architecture we tested was the same as the QuatE-1 architecture, but with the real-valued, fully connected backend replace by a 4D PHM layer (QPHM). We call this ``QuatE-2''.
Experiments with QuatE-2 tell us about the effects of using a representationally coherent backend.

The third architecture we tested replaced the 7x7 real-valued, convolutional stem with a 7x7 quaternion-valued convolutional stem. We also used the bottleneck block shown in Fig.~\ref{fig:proposedWithAttention}(d) instead of that in Fig.~\ref{fig:proposedWithAttention}(c). In summary, this architecture used a convolutional stem, bottleneck blocks like those in Fig.~\ref{fig:proposedWithAttention}(d), and the 4D PHM layer.
That is, all possible layers were replaced with representationally coherent versions. We call this 
``RepAA''.

\section{Experiment-1}
\label{sec:exp-1}
We compare four models on a subset of the ImageNet dataset, called ImageNet300k, which we created (explained below).
We use this dataset because we did not have the computing power to conduct simulations using 
the full ImageNet dataset \cite{deng2009imagenet}.

\subsection{Method}

The models we compare are: the standard convolution-based ResNet \cite{heresnet2016}, 
the QCNN \cite{gaudet2018}, 
the axial-ResNet \cite{wang2020axialdeeplab},
and our novel quaternion-enhanced axial-ResNet.
The main objective is to see if the representations
generated by the quaternion front end improve classification performance
of the axial attention model.
The axial-ResNet and QuatE-1 axial-ResNet models are depicted in Table~\ref{tab_archiTable250}.
We used a 26-layer version with the block multipliers ``[1, 2, 4, 1]'' and a 50-layer version with the block multipliers ``[3, 4, 6, 3]'' of the models.
Table~\ref{tab_archiTable250} shows the 50-layer versions.
The bracketed expressions show the bottleneck blocks with the operations used and
the number of output channels for 
each stage.
If the quaternion modules are counted then the layer counts for our model are 33 layers and 66 layers, respectively.
We count the two-1D-layer axial-attention module as one layer because two 1D layers are equivalent to one 2D layer. 

Because of hardware limitations, our experiment conducts image classification on a subset of the ImageNet dataset \cite{russakovsky2015imagenet}. Specifically, it took over two hours to train one of the axial-ResNet models for one epoch.
This smaller dataset which requires lower computing resources, called ImageNet300k,  uses the same 1,000 image categories as 
the original ImageNet \cite{deng2009imagenet}. 
The full ImageNet dataset has 1.28 million training images and 50,000 validation images. Our smaller dataset, which we call ImageNet300k, is sampled from the full ImageNet by taking 
the first 300 images for each category contained in the original dataset, yielding 300,000 training images. There are also 50,000 validation images with 50 images per category. These are from the original dataset. Although the training dataset is smaller, it still
allowed us to train our 50-layer networks without overfitting. 
Overfitting was assessed by examining performance on the validation dataset in comparison to the training set. The validation dataset was the same as that used in the original ImageNet dataset.

\begin{table*}
\centering
\begin{tabular}{|l|c|c|r|c|r|} \hline
\shortstack{Architecture} & Layers & \shortstack{Training\\Top-1 Acc.} & Params & \shortstack{Validation\\Top-1 Acc.} & Inference time\\ 
\hline
ResNet \cite{heresnet2016}   &26&57.0&13.6M&45.48&8.86 msec\\
Quaternion ResNet \cite{gaudet2018} &26&64.1&15.1M&50.09&25.32 msec\\
Axial-Attention \cite{wang2020axialdeeplab}    &26&61.0  & 5.7M&54.79&27.94 msec\\
QuatE-1 Axial-Attention (ours)&33&78.2& 6.0M&62.30&31.75 msec\\\hline
ResNet \cite{heresnet2016} &50&65.8&25.5M&50.92&14.70 msec\\
Quaternion ResNet \cite{gaudet2018}  &50&73.4&27.6M&49.69&50.01 msec\\
Axial-Attention \cite{wang2020axialdeeplab}     &50&63.6&11.5M&55.57&52.35 msec\\
QuatE-1 Axial-Attention (ours)&66&72.6&11.9M&59.71&58.41 msec\\\hline
\end{tabular} 
\caption{Image classification performance on the ImageNet300k dataset for 26 and 50-layer architectures. We include Top-1 training and validation accuracies.}
\label{tab_resultTable}
\end{table*}

All network models (Experiment-1, 2, and 3) were trained using the same stochastic gradient descent optimizer and hyperparameters. All networks were trained for 150 epochs using stochastic gradient descent optimization with momentum set to 
0.9 and the learning rate which is warmed up linearly from $\epsilon$ near zero to 0.1 for 10 epochs 
\cite{wang2020axialdeeplab}. The learning rate was then cut by a factor of 10 at epochs 20, 40, and 70.
We adopt the same training protocol as \cite{wang2020axialdeeplab} except for batch size. 
Our batch size is limited to ten because of memory limitations. 
Also, we used batch normalization with 0.00009 weight decay. 
For attention models, the number of attention heads is fixed to eight in all attention layers \cite{wang2020axialdeeplab}.

\subsection{Results Analysis}
The main results are seen in Table~\ref{tab_resultTable}.
The top half of the table shows the results for the 26-layer models (33-layers for the
QuatE-1 model).
The bottom half shows the results for the 50-layer models (66-layers for the QuatE-1 model).
These are the parameter count, the inference time, and the percent accuracy results for a single simulation run for each model. Both of the axial-ResNets use far fewer parameters
than the convolution-based ResNets.

The most important comparison is between the axial-ResNet and the QuatE-1 axial-ResNet because this directly shows the effect of the quaternion-generated interwoven/interlinked representations. There are two such comparisons, one for the 26/33 layer models and the other for 
the 50/66 layer models. In both cases, the quaternion enhanced versions of axial-ResNet produced higher classification accuracy performance.
This was true for both the training and validation data.
This supports our main hypothesis that quaternion modules can produce more usable interlinked/interwoven representations. Of course, there is the alternative explanation that the quaternion enhanced axial-ResNet
had more layers and this is the reason for the better performance.
This is addressed in the next section.

Another result is that both of the axial-ResNet architectures provide higher performance for
the validation set
than either of the convolution-based ResNets.
This is true for both the 26-layer (33-layer) and 50-layer (66-layer) versions.

Finally, it is surprising that the 33-layer quaternion-enhanced axial-ResNet gives better classification accuracy performance than the 66-layer version. This is true for both the training and validation data. We have no explanation for this. If this were only true for the validation data, then overfitting would be a possible explanation. This is addressed in the Experiment-3 section.

\section{Experiment-2}
\label{sec:exp-2}
In the above experiment at Table~\ref{tab_resultTable}, adding the quaternion front-end to the axial attention module
increased the number of layers from 26 to 33 for the small network and from 50 to 66 for the large network. This offered an alternative explanation for the improved classification results. The accuracy improvement might have been caused by the increased number of layers
and not the added quaternion bank. This analysis tests the alternative explanation.

\subsection{Method}
This experiment increases the layer count of the 
baseline networks from 26 to 35 layers. This was done by using a block multiplier of [2, 3, 4, 2] for these models.
Although this did not give us exactly 33 layers to exactly match the QuatE-1 axial-attention model, it preserved the bottleneck structure of the original design and enhances comparability.

\begin{table*}
\centering
\begin{tabular}{|l|c|c|r|c|} \hline
Architecture & Layers & Training Top-1 Acc. & Params & Validation Top-1 Acc.\\ 
\hline
ResNet \cite{heresnet2016} &35&63.8&18.5M&48.99\\
Quaternion ResNet \cite{gaudet2018} &35&70.9&20.5M&48.11\\
Axial-Attention \cite{wang2020axialdeeplab} &35&73.6& 8.4M&60.49\\\hline
\end{tabular}
\caption{Image classification performance on the ImageNet300k dataset for ResNet-35 architectures. 
We include Top-1 training and validation accuracies.}
\label{tab_resultTable2}
\end{table*}

\subsection{Experimental Results}
Table~\ref{tab_resultTable2} shows the parameter count and accuracy for these 35-layer models.
The most important comparison is the 35-layer Axial-Attention Network in Table~\ref{tab_resultTable2}
with the 33-layer QuatE-1  Network in Table~\ref{tab_resultTable}.
Now the layer count for the non-quaternion version is slightly larger than that for the
quaternion version but the quaternion version still gives better performance
on both the training and validation accuracy.
The increased layer count did improve the performance of the 35-layer Axial Attention Network
but not enough to surpass the quaternion version. Thus, we conclude that the quaternion frontend has more impact than the layer count.

\section{Experiment-3}
\label{sec:exp-3}
By looking at Tables \ref{tab_resultTable} and \ref{tab_resultTable2}, we are surprised that the 33-layer QuatE-1 model gives better classification accuracy performance than the 66-layer version. We did not have any explanation why are these happening? To handle these situation, we performed another experiment named RepAA networks on the same ImageNet300k dataset.

\subsection{Method}
This experiment used the RepAA model on ImageNet300k dataset. The performances of QuatE axial-ResNets with QPHM in the back-end, and RepAA networks are compared in Tables \ref{tab_resultTable}, \ref{tab_resultTable2}, and \ref{tab_resultTable3}. 
The architectural details of RepAA network (50-layer version) are shown in Table~\ref{tab_archiTable250}. In this experiment, we ran our models for 26-layer, 35-layer, and 50-layer architectures with the same multipliers and 
we trained RepAA networks using the same hyperparameters as like in Experiment-1 and 2. 

\begin{table*}
\centering
\begin{tabular}{|l|c|c|r|c|} \hline
Architecture & Layers & \shortstack{Training \\ Top-1 Acc.} & Params & \shortstack{Validation\\ Top-1 Acc.} \\ 
\hline
ResNet \cite{heresnet2016} &26&57.0&13.6M&45.48\\
Quaternion ResNet \cite{gaudet2018} &26&64.1&15.1M&50.09\\
Axial-Attention \cite{wang2020axialdeeplab} &26&61.0  & 5.7M & 54.79\\
RepAA network (ours) &26& 73.0 & 3.7M & 60.70\\
\hline
ResNet \cite{heresnet2016} &35&63.8&18.5M&48.99\\
Quaternion ResNet \cite{gaudet2018} &35&70.9&20.5M&48.11\\
Axial-Attention \cite{wang2020axialdeeplab} &35&73.6& 8.4M&60.49\\
QuatE-1 Axial-Attention (ours) &33&78.2&6.0M&62.30\\
QuatE-2 Axial-Attention with QPHM (ours) &33&75.5&5.3M&61.27\\
RepAA network (ours) &35&72.0&4.6M&62.03\\
\hline
ResNet \cite{heresnet2016} &50&65.8&25.5M&50.92\\
Quaternion ResNet \cite{gaudet2018} &50&73.4&27.6M&49.69\\
Axial-Attention \cite{wang2020axialdeeplab} &50&63.6&11.5M&55.7\\
QuatE-1 Axial-Attention (ours) &66&72.6&11.9M&59.71\\
QuatE-2 Axial-Attention with QPHM (ours) &66&77.8&11.1M&62.46\\
RepAA network (ours) &50& 73.5 &6.7M& 62.49\\
\hline
\end{tabular}
\caption{Image classification performance on the ImageNet300k dataset for 26, 35 and 50-layer architectures. 
We include Top-1 training and validation accuracies. ``QuatE'' and ``RepAA'' stands for quaternion enhanced and representational axial-attention.}
\label{tab_resultTable3}
\end{table*}

\subsection{Experimental Results}
We next investigate the use of our final proposed method, the representational axial-attention network, to see how it performs on the task of ImageNet300k dataset. Table~\ref{tab_resultTable3} shows our overall experiment results for the 26, 35, and 50-layer architectures. Among them, our final proposed method performs better in accuracy and the trainable parameters. 
\begin{figure*}
    \centering
    \begin{subfigure}[b]{0.45\textwidth}
        \centering
        \includegraphics[width=\textwidth]{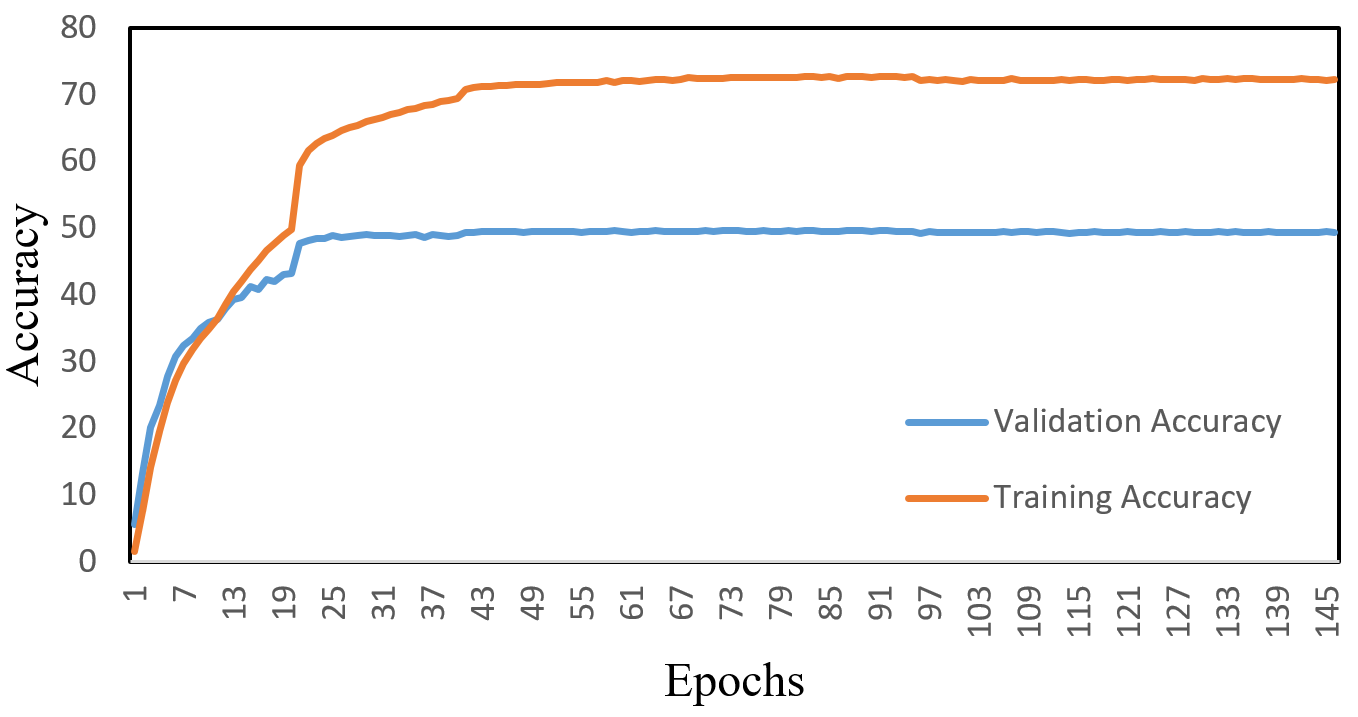}
        \caption{Original quaternion network}
    \end{subfigure}
    \hfill
    \begin{subfigure}[b]{0.45\textwidth}
        \centering
        \includegraphics[width=\textwidth]{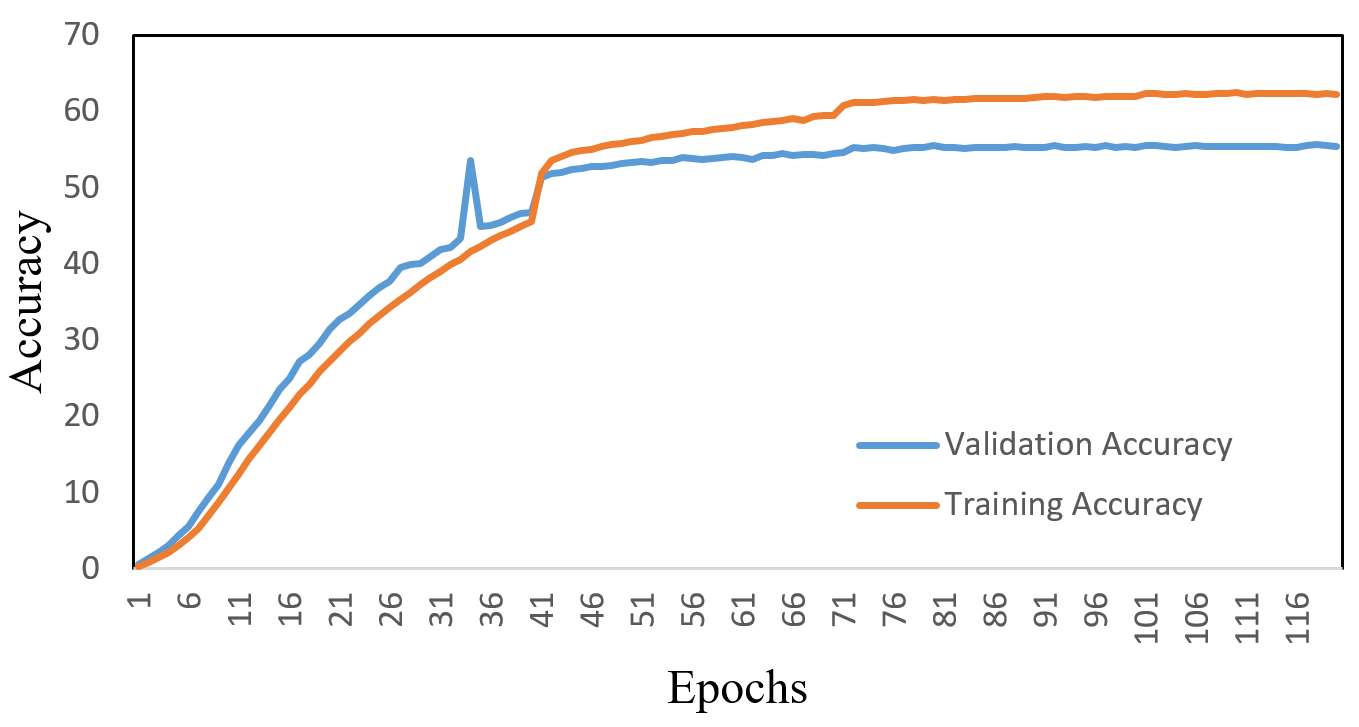}
        \caption{Original axial-attention network}
    \end{subfigure}
    \hfill
    \begin{subfigure}[b]{0.45\textwidth}
        \centering
        \includegraphics[width=\textwidth]{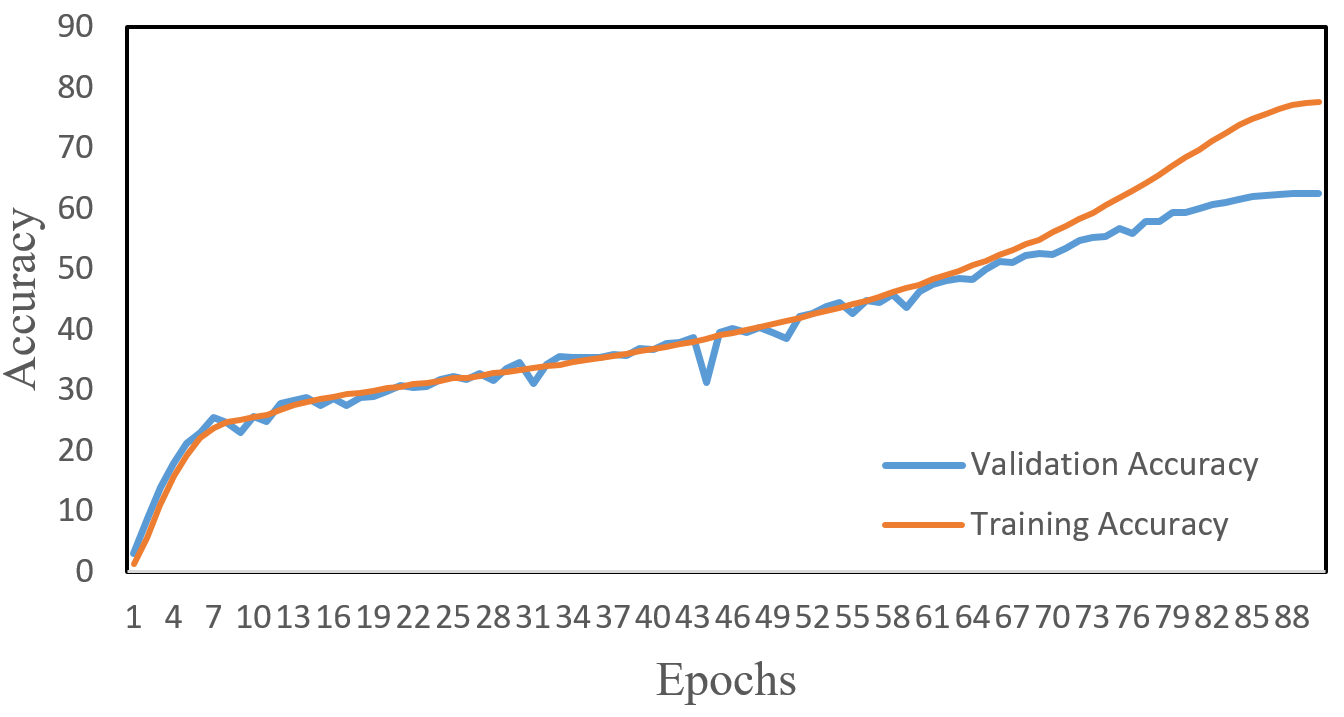}
        \caption{Proposed QuatE-2 axial-attention network}
    \end{subfigure}
    \hfill
    \begin{subfigure}[b]{0.45\textwidth}
        \centering
        \includegraphics[width=\textwidth]{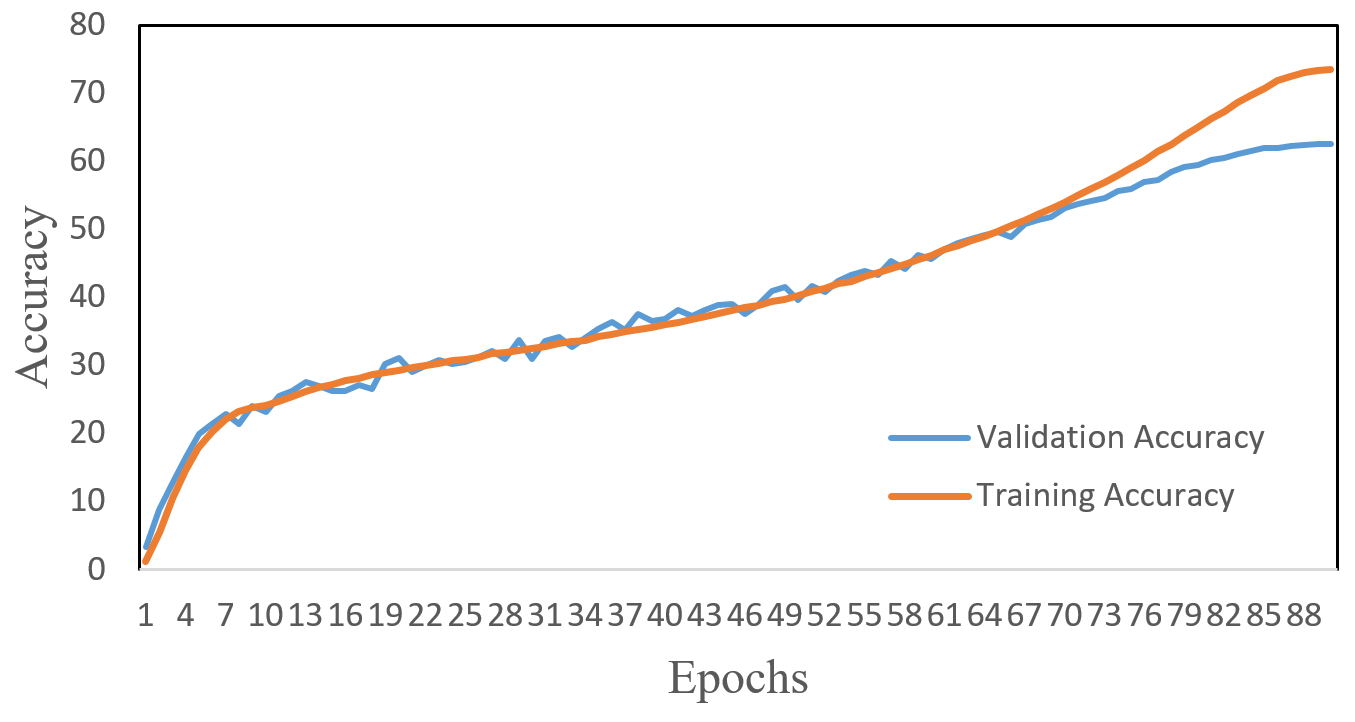}
        \caption{Proposed RepAA method}
    \end{subfigure}
    \hfill 
    \caption{Training and validation performance comparison of original quaternion ResNet, original axial-attention network, QuatE-2 axial ResNet with QPHM (ours), and our proposed RepAA method (50-layers version) to assess the overfitting for ImageNet300k dataset.}
    \label{fig:AccuracyComparison}
\end{figure*}
\begin{figure*}
    \centering
    \begin{subfigure}[b]{0.48\textwidth}
        \centering
        \includegraphics[width=\textwidth,height=130pt]{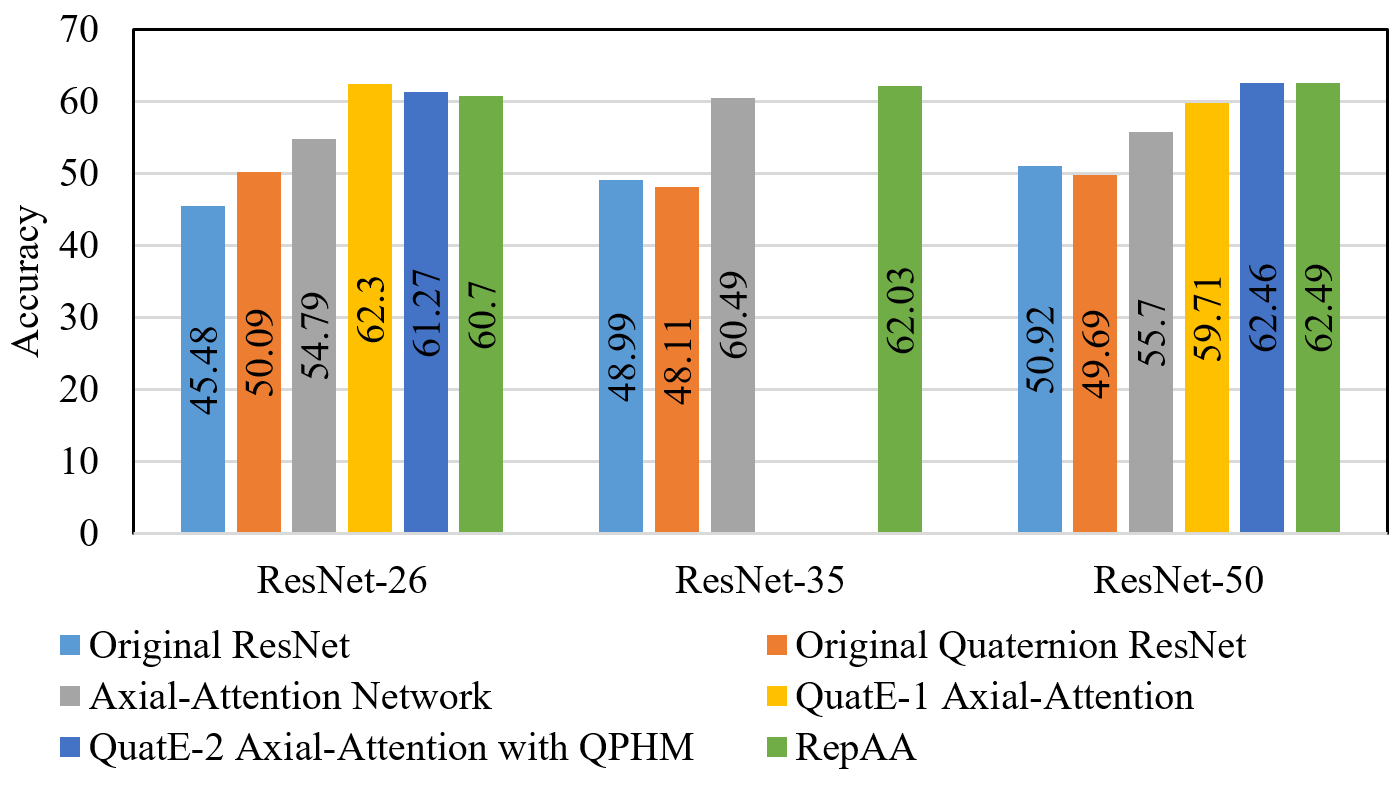}
        \caption{Validation Performance comparison}
    \end{subfigure} 
    \hfill
    \begin{subfigure}[b]{0.48\textwidth}
        \centering
        \includegraphics[width=\textwidth,height=130pt]{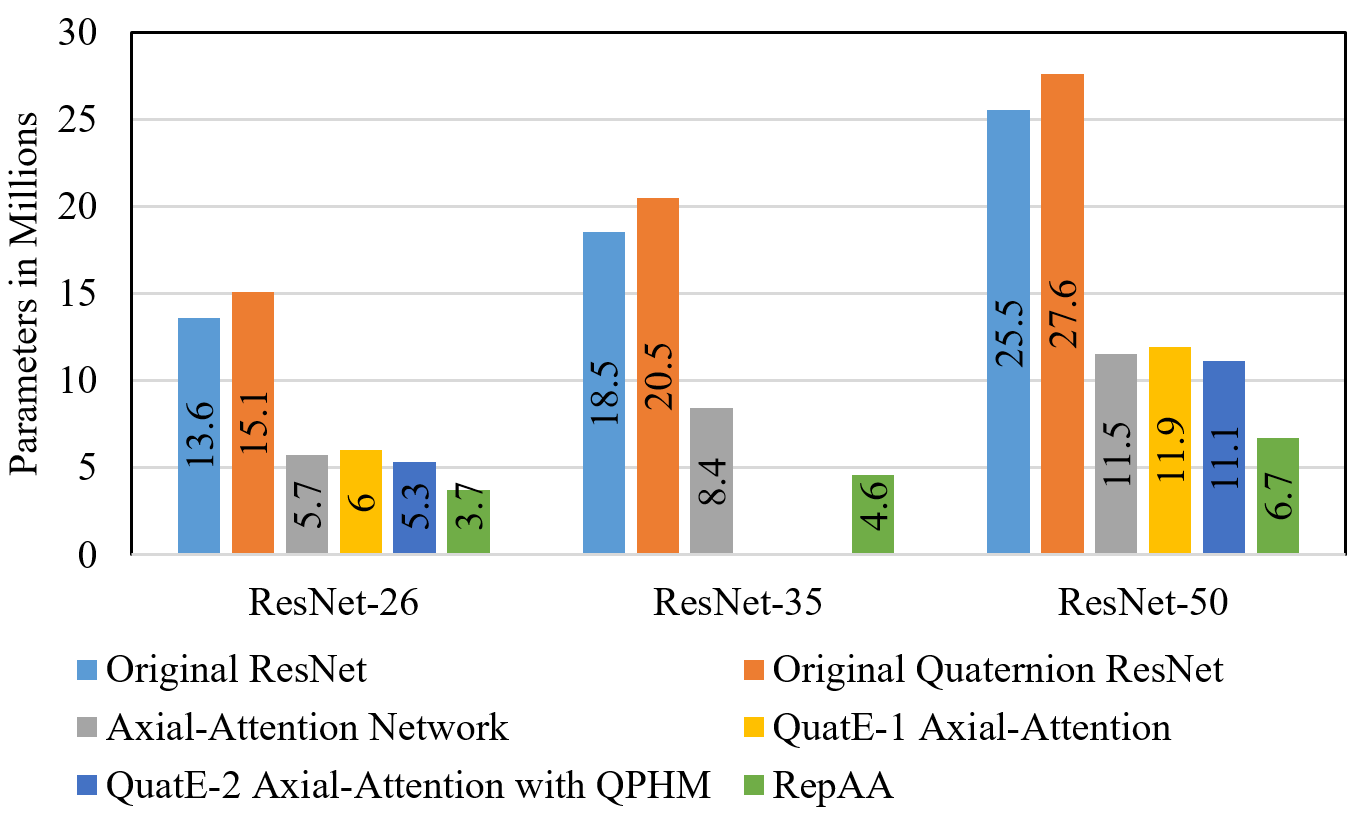}
        \caption{Trainable parameter comparison}
    \end{subfigure}
    \caption{Validation performance accuracy (in percent) and trainable parameter (in millions) comparison of ResNet, original quaternion ResNet, original axial-attention network, QuatE-1 axial ResNet, QuatE-2 Axial with QPHM, and RepAA.}
    \label{fig:PerformanceAndParameterComparison}
\end{figure*}

Figure \ref{fig:AccuracyComparison} assesses the overfitting problem by examining the performance on the validation dataset in comparison to the training dataset and yields no overfitting for ImageNet300k dataset. 
The most important comparisons are between QuatE axial-ResNet \& our proposed RepAA model, and  axial-ResNet \& RepAA model as these directly show the effect of removing extra $1\times 1$ quaternion layer from QuatE axial-ResNet and applying representational effect throughout the attention network. In both cases,
the RepAA networks produced higher classification accuracy with far fewer parameters. In general, our proposed method outperforms then the other networks with \textbf{1.5 times}, \textbf{1.8 times} and \textbf{1.7 times} fewer parameters in compared to original axial ResNet for 26, 35, and 50-layer networks, respectively. 

Although, we ran our model for only 90 epochs, which is less than the other models, but it's performance was unchanged for 150 epochs. Unlike QuatE axial-ResNets, the validation performance for 35-layers is higher than the 26-layers, and 50-layers is outperformed than the others. This supports our main hypothesis, quaternion modules can produce more usable interlinked/interwoven representations.

\section{Conclusions and Future Work}
Recall that we replaced traditional modules with representationally coherent modules
in the stem, the bottleneck blocks, and the fully connected backend.
We found that all of these novel modifications improved accuracy to varying degrees when
trained and tested on the ImageNet300k dataset.
Our baseline networks for comparison were the real-valued ResNet, the QCNNs, and the Axial Attention ResNet.

Our results may be significant because the improvement was observed when any part of
the network was modified to use representational coherence.
This suggests that
there is a promise that this technique may be generally useful in improving
classification accuracy for a large class
of networks. This work was limited to the ImageNet300k dataset due to machine limitations and wasn't able to evaluate low-resolution datasets like CIFAR benchmarks as attention models require a large number of high-resolution data to handle the underfitting problem.
Further work may be directed at examining other architectures to see if this
claim is true.
Additional further work should be directed toward testing whether these results
hold up when tested on other datasets.

Finally, vector maps and PHM layers offer more fine-grained control over weight
sharing than quaternions. For quaternions, 
four weights
are distributed over 16 slots, diminishing the weight count ratio to 25\%.
Since vector map and PHM operations are not constrained to four dimensions,
other weight count ratios can be tested.

{\small
\bibliographystyle{ieee_fullname}
\bibliography{egbib}
}

\end{document}